\documentclass[letterpaper, 10 pt, conference]{ieeeconf}

\IEEEoverridecommandlockouts
\usepackage{multicol}
\usepackage{multirow}
\usepackage[bookmarks=true]{hyperref}

\usepackage[utf8]{inputenc} 
\usepackage[T1]{fontenc}    
\usepackage{url}            
\usepackage{booktabs}       
\usepackage{amsfonts}       
\usepackage{nicefrac}       
\usepackage{microtype}      
\usepackage{amsmath,amssymb}
\usepackage{mathtools}
\usepackage{array}

\usepackage{algorithm}

\usepackage{algorithm}
\usepackage[noend]{algpseudocode}
\usepackage{subcaption}
\usepackage[table, dvipsnames]{xcolor}
\usepackage{adjustbox}
\usepackage{wrapfig}
\usepackage{mdframed}
\usepackage{makecell}
\usepackage{siunitx}
\usepackage{flushend}
\usepackage{todonotes}
\usepackage{bm}
\usepackage{multirow}

\makeatletter
\newcommand{\vast}{\bBigg@{4}}
\newcommand{\Vast}{\bBigg@{5}}
\makeatother
\usepackage{pgfplots}
\usepackage{tikz}
\usepgfplotslibrary{groupplots}
\usepackage{siunitx}

\newcommand{\sn}[1]{\num[round-precision=3,round-mode=figures,scientific-notation=true]{#1}}
\makeatletter
\def\hiddenfootnote{\gdef\@thefnmark{}\@footnotetext}
\makeatother

\newcolumntype{L}[1]{>{\raggedright\arraybackslash}p{#1}}

\title{Learning and Blending Robot Hugging Behaviors in Time and Space}

\author{Michael Drolet$^{1}$, Joseph Campbell$^{2}$, and Heni Ben Amor$^{1}$\\
$^{1}$
Arizona State University\\
$^{2}$
Carnegie Mellon University\\
        {\tt\small \{mdrolet, hbenamor\}@asu.edu, jcampbell@cmu.edu}%
}

\begin{document}

\maketitle

\begin{abstract}
We introduce an imitation learning-based physical human-robot interaction algorithm capable of predicting appropriate robot responses in complex interactions involving a superposition of multiple interactions.
Our proposed algorithm, Blending Bayesian Interaction Primitives (B-BIP) allows us to achieve responsive interactions in complex hugging scenarios, capable of reciprocating and adapting to a hug's motion and timing.
We show that this algorithm is a generalization of prior work, for which the original formulation reduces to the particular case of a single interaction, and evaluate our method through both an extensive user study and empirical experiments.
Our algorithm yields significantly better quantitative prediction error and more-favorable participant responses with respect to accuracy, responsiveness, and timing, when compared to existing state-of-the-art methods. \\
\end{abstract}

\IEEEpeerreviewmaketitle

\section{Introduction}
\label{sec:introduction}
A hug is a natural embrace and one of the most common forms of social interaction in humans and animals.
Beyond a simple salutation, it is an effective means to communicate affection and emotional support \cite{holt2008influence}, and studies have shown hugging to cause physiological responses, thereby resulting in cardiovascular and mental health benefits \cite{light2005more, grewen2003warm, cohen2015does}. With the advent of social robotics, e.g., robots in malls, homes, and theme parks, there is an increasing need for methods that can produce responsive and convincing hugging motions in anthropomorphic agents. However, despite its seemingly simple appearance, hugging is a nuanced and complex process of physical coordination in both time and space.

When execution fails, we are often left feeling uncomfortable, awkward, or embarrassed -- the opposite of its intended effect. In turn, implementing such behaviors in robots is an extremely challenging task and often circumvented in experimental hugging robots by not reciprocating hugs~\cite{yamazaki2016intimacy, sumioka2013huggable} or executing non-adaptive, pre-defined motions~\cite{block2019softness, shiomi2017robot, hedayati2019hugbot}. Adaptive hugs require a robot to anticipate the type of hug performed, the current temporal progress, and the upcoming motion. Consequently, it has to generate accurate motor behavior to produce a synchronized motion with the human partner. One challenge in this regard is that hugs can be initiated anytime. Hence, the starting point is not predetermined nor easily identifiable. Another challenge is that hugs are typically fast movements with a duration of only a few seconds. Therefore, we need algorithms that allow robots to repeatedly (a) replan their motions in response to (b) the predicted behavior of the human partner. Finally, due to cultural and personal preferences, there may be a number of variations and styles of hugs.

\begin{figure}[t]
    \centering
    \includegraphics[width=\linewidth]{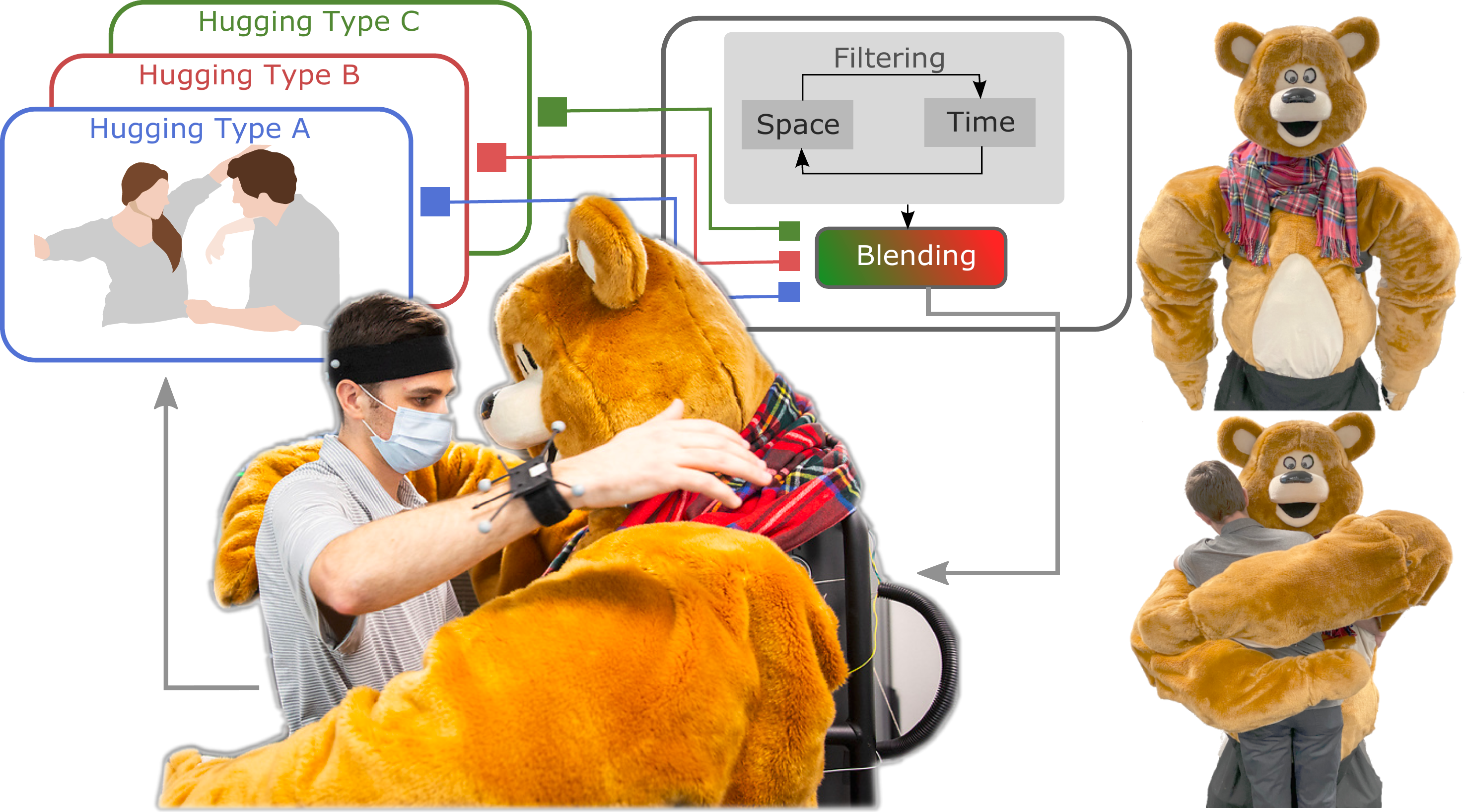}
    \caption{Robot dressed as a plush bear hugs a human partner.}
    \label{fig:exp_setup}
\end{figure}

We seek to address the above challenges by introducing a human-robot interaction framework that maintains a list of possible interactions and fluidly transitions between them in response to the interaction partner.
Traditionally, HRI methods require us first to classify which interaction is occurring, then proceed with that classification for the remainder of the interaction~\cite{amor2014interaction, ewerton2015learning}, or to re-classify the interaction at discrete time intervals.
However, given the dynamic nature of human interactions, this leads to stilted HRI in which the robot is incapable of smoothly transitioning between discrete actions or even accounting for an interaction that is a blend of multiple actions.

We introduce a generalized version of Bayesian Interaction Primitives~\cite{campbell2017bayesian} in which interaction may consist of multiple sub-actions.
In this context, the original formulation may be considered a particular case of our proposed general form.
At each time step, we assess the likelihood of a set of possible interactions based on observations of the human partner.
The model associated with each possible interaction is then updated based on the observations, with the magnitude of the update weighted by the likelihood.
The intuition here is straightforward: if we are observing a seemingly unlikely interaction, then we do not want to update the model significantly because the observation is unlikely to have been generated from the model. It would only serve to distort it.
Aside from being able to responsively transition from one interaction to another at any point in time, our approach has two subtle advantages: a) by updating all likely interactions at every time step, we avoid a sudden discontinuous transition between discrete interactions when switching occurs, and b) it is possible that interaction is a blend of multiple discrete interactions.

We propose the following contributions in this work:
\begin{itemize}
    \item A generalized form of Bayesian Interaction Primitives, which removes the restriction that a single interaction consists of only a single action.
    \item A probabilistic formulation in which we detect action transitions and update a state approximation.
    \item An empirical study in which we demonstrate that our formulation can successfully detect and transition between three discrete sub-actions in a physical hugging scenario and compare it to baseline methods.
\end{itemize}

\section{Related Work}

\subsection{Social and Physical Human-Robot Interaction}

Intimate, social pHRI, such as hugging, has been found to have positive effects on the human emotional state~\cite{wada2004effects,rus2015design,shiomi2017robot,sumioka2013huggable}. Subsequently, robotic hugging has been an area of interest as it would be beneficial if robots could confer the same positive emotional benefits as humans. While passive, non-reciprocating huggable robots have been shown to yield emotional, mental, and physical health benefits~\cite{disalvo2003hug, sumioka2013huggable, yamazaki2016intimacy}, robots which actively reciprocate hugs have been found to lead to greater interaction duration~\cite{shiomi2021robot} and are well received if the robot is responsive and comfortable to hug~\cite{block2019softness}.

However, given the challenges involved in developing adaptive hugging algorithms, many robots capable of reciprocating hugs use pre-defined motions in a "one size fits all" setup~\cite{shiomi2017robot, gaitan2021physical, hedayati2019hugbot}. Sometimes, limited adaptability is achieved by controlling when the hug is initiated and ended based on visual and haptic feedback~\cite{block2021six}.
More recently, recognition and reciprocation of intra-hug gestures have been investigated~\cite{block2022arms}, leading to another layer of responsiveness. However, the timing and motion of the hug itself are still largely independent of the user once initiated. It is this latter adaptability for which we propose a solution in this work.

\begin{figure*}
\centering
\includegraphics[width=0.99\textwidth]{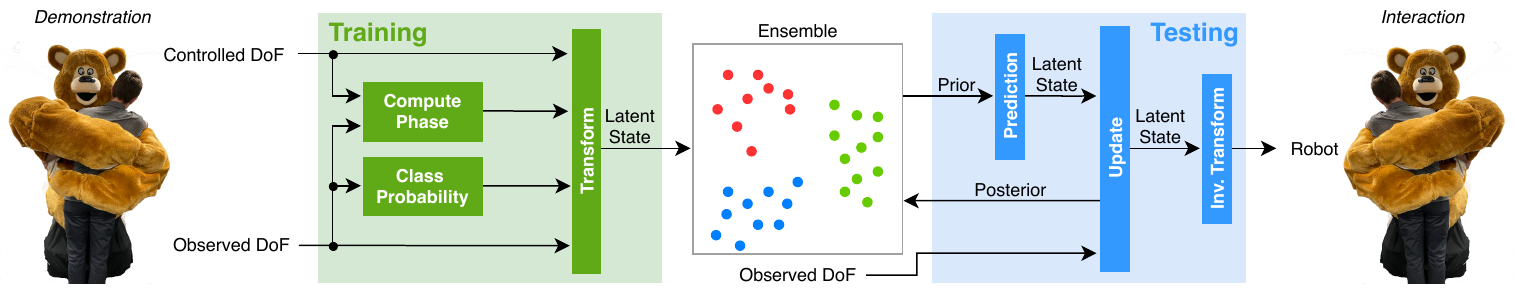}
\caption{An overview of B-BIP. Top: training demonstrations (left) are decomposed into a latent space (middle) and transformed into an ensemble of samples (right). Bottom: observations are collected during a live interaction (left) which is used to perform filtering with the learned ensemble (middle) and produce a response trajectory (right).}
\label{fig:overview}
\end{figure*}

\subsection{Learning Interactions with Multiple Sub-Actions}

Methods to combine multiple motion actions, i.e., primitives, into complex movements have been extensively studied, both in terms of sequences~\cite{kulvicius2011joining, kober2015learning} and superpositions~\cite{williams2007modelling, paraschos2013probabilistic}.
However, such approaches have primarily been limited to movement generation for a robot acting independently without coordinating its behavior with a human partner.

Methods targeting human-robot interaction scenarios are often based on graphical models.
For example, Tanaka et al.~\cite{tanaka2012motion} developed a Markov model-based approach to determining which discretized state space region the human partner would be in so the robot could act appropriately.
Koppula et al.~\cite{koppula2015anticipating} demonstrate a method based on conditional random fields which can anticipate future human behavior and classify the specific sub-actions.
Graphical models are employed by Hawkins et al.~\cite{hawkins2013probabilistic} to develop a method in which a robot plans its actions to reduce an expected cost associated with a human's predicted sub-action timings.

In the method most similar to ours, Ewerton et al.~\cite{ewerton2015learning} present a method to accommodate multiple interaction primitives but do not support a blend or superposition of primitives. Additionally, this method assumes the phase is known/fixed (where observation comes from the last time step of interaction), and then forward control is applied to execute the trajectory. However, we address the problem more generally, in that phase is treated as a random variable. Therefore, our algorithm brings novelty in the sense that we can estimate the correct primitive at an arbitrary phase, which does not need to be provided beforehand; and due to the online, closed-loop nature of the inference algorithm, we can blend between interactions in real-time.
In contrast to prior work, we present a fully probabilistic model at the motion planning level, which infers both the current action and the motion and temporal timing into the future.

\section{Methodology}
In this work, we model a single physical human-robot interaction as a Bayesian Interaction Primitive~\cite{campbell2017bayesian}.
Each primitive explicitly models the relationship between the measured degrees of freedom from a set of training demonstrations, which is then used as a prior for inference during interactions, as depicted in Fig.~\ref{fig:overview} in the \emph{Training} block.

The goal is for a robot to interact with a human partner in real-time and infer their (a) next actions, and (b) appropriate robot response using this prior knowledge and a sequence of observations of the human, shown in the \emph{Testing} block of Fig.~\ref{fig:overview}.
We begin by describing this model in detail, then introduce an improved version, Blending BIP (B-BIP), which allows us to expand a single primitive to encompass multiple interactions while enabling transitions between them.

\subsection{Preliminaries: Bayesian Interaction Primitives}

We first train the primitive using a set of demonstrations of the desired interaction.
Each demonstration consists of \emph{observed} and \emph{controlled} degrees of freedom, which we model as the matrix $\bm{Y} \in \mathbb{R}^{D \times T}$ where $D = |D_c| + |D_o|$ denotes the total number of degrees of freedom (DoFs) in the interaction (having the sets of DoF indices $D_c$ from the controlled agent and $D_o$ DoFs from the observed agent) and $T$ samples.

These demonstrations are then transformed to a time invariant latent space $\boldsymbol{w}$ such that $y_t^d = h^d(\phi(t), \boldsymbol{w}^d) = \Phi_{\phi(t)}^{\intercal} \boldsymbol{w}^d + \epsilon_y$, where $\Phi_{\phi(t)} \in \mathbb{R}^{1\times B}$ is a row vector of $B^d$ basis functions, $\boldsymbol{w}^d \in \mathbb{R}^{B \times 1}$, and $\epsilon_y$ is i.i.d. Gaussian noise.
We use the standard Gaussian basis function in this work, although others may be selected if appropriate for the task domain.
The full latent state representation for a demonstration $\boldsymbol{w}$ is obtained by concatenating each degree of freedom together, and may be solved for using standard optimization techniques such as least squares.
As is standard~\cite{campbell2017bayesian}, the basis functions are dependent on a relative time measure referred to as phase $\phi(t)$.

During inference, we wish to estimate the latent state $\boldsymbol{w}$ from which the inferred controlled DoFs can be retrieved, however, it is also necessary to localize both the phase, $\phi$, and phase velocity, $\dot{\phi}$, in order to accurately perform inference on demonstrations done at different speeds. 
Therefore, we augment the state representation to be $\bm{s}_t = [\bm{\phi}_t, \bm{\dot{\phi}}_t, \bm{w}_t]$.
Given a sequence of measurements, $\bm{Y}_{1:t}$, of all modalities, we have the following probabilistic formulation
\begin{equation}
\label{eq:bip_general}
p(\boldsymbol{s}_t | \boldsymbol{Y}_{1:t}, \boldsymbol{s}_{0}) \propto p(\boldsymbol{y}_{t} | \boldsymbol{s}_t) p(\boldsymbol{s}_t | \boldsymbol{Y}_{1:t-1}, \boldsymbol{s}_{0}).
\end{equation}
As in~\cite{campbell2019probabilistic}, this posterior density is approximated using a Monte Carlo method in which an ensemble of samples are used to predict the next measurement,
\vspace{-1mm}
\begin{align}
\boldsymbol{x}^j_{t|t-1} &=
g(\boldsymbol{x}^j_{t-1|t-1})
+
\mathcal{N}
\left(0, \boldsymbol{Q}\right).
\label{eq:state_prediction}
\end{align}
The predicted ensemble is updated based on the error obtained from the actual measurement using a gain coefficient $\boldsymbol{K}$, the equations for which are omitted due to space constraints. The interested reader can refer to the standard Ensemble Kalman Filter (EnKF)~\cite{evensen2003ensemble} algorithm for more information.
\begin{align}
\boldsymbol{x}^j_{t|t} &= \boldsymbol{x}^j_{t|t-1} + \boldsymbol{K} (\boldsymbol{\tilde{y}}_{t} - h(\boldsymbol{x}^j_{t|t-1})).
\label{eq:measurement_update}
\end{align}

The set of training demonstrations that we start with are used to directly initialize the ensemble members, such that $E$ demonstrations yields $E$ ensemble members.

\subsection{Modeling Multiple Primitives}
\label{sec:method_multiple}

The BIP framework as described above only supports modeling a single interaction at a time.
In order to extend this model to a set of interactions, we first present a probabilistic formulation for what this entails.
Suppose we have an ensemble $\boldsymbol{X}$ in which each of the $E$ ensemble members belongs to a class $c \in \mathcal{C}$, in which the set of classes $\mathcal{C}$ represents different discrete interactions.
This allows us to partition $\boldsymbol{X}$ into $|\mathcal{C}|$ sub-ensembles, such that $\boldsymbol{X} = \boldsymbol{X}^1 \cup \boldsymbol{X}^2 \cup \dots \cup \boldsymbol{X}^{|\mathcal{C}|}$ where a sub-ensemble $\boldsymbol{X}^c$ contains $E^c$ members such that $E = \sum_{c \in \mathcal{C}} E^c$.
We define $C$ as a random variable over the set $\mathcal{C}$ which indicates the class of an interaction.
Each sub-ensemble $\boldsymbol{X}^c$ is a Monte Carlo approximation of the probability distribution for the $c$-th interaction, $p(\boldsymbol{s}_t | \boldsymbol{Y}_{1:t}, \boldsymbol{s}_0, C = c)$, for which we can marginalize out $C$ to re-obtain the full posterior distribution:
\begin{align}
    p(\boldsymbol{s}_t |& \boldsymbol{Y}_{1:t}, \boldsymbol{s}_0) = \nonumber \\
    & \sum_{c \in \mathcal{C}} p(\boldsymbol{s}_t | \boldsymbol{Y}_{1:t}, \boldsymbol{s}_0, C = c) p(C = c | \boldsymbol{Y}_{1:t}, \boldsymbol{s}_0).
    \label{eq:prob_classes}
\end{align}

The association of each ensemble member to a class $c$ is static and defined in the prior distribution $\boldsymbol{s}_0$, as demonstrations must be initially collected for each individual interaction and hence we have a mapping from demonstrations to classes.
This allows us to calculate the posterior for a specific class, $p(\boldsymbol{s}_t | \boldsymbol{Y}_{1:t}, \boldsymbol{s}_0, C = c)$, in a similar manner as Eqs.~\ref{eq:state_prediction}-\ref{eq:measurement_update} but limited to the ensemble members $\boldsymbol{x} \in \boldsymbol{X}^c$; this is covered in Sec.~\ref{sec:transition}.
We do not restrict ourselves to the case that $C$ is fixed to a single value $c$; an interaction may transition between multiple classes over time.
Therefore, the interaction scenarios examined in previous works are special cases of this formulation and only take on one class value.

\subsection{Interaction Detection}
In this work, we peform Reduced-Rank Linear Discriminant Analysis (LDA) for computing the probability of the interaction class, given the current observations of the human. For each interaction class, $c \in \mathcal{C}$, let $\bm{Y}^{(c)}_i \in \mathbb{R}^{D \times T}$ represent the $i$'th demonstration from the set of training demonstrations of the corresponding class.
Here, $1 \leq i \leq N^{(c)}$, with $N^{(c)}$ denoting the total number of demonstrations for class $c$. 
Let the within-class demonstration matrix be defined by $\bm{M}^{(c)} = [ (\bm{Y}_1^{(c)})_{D_o, :} \hspace{1mm}, \hspace{1mm} \dots \hspace{1mm} , \hspace{1mm} (\bm{Y}_{N^{(c)}}^{(c)})_{D_o, :} ]$ and the between-class demonstration matrix be defined by $\bm{M} = \left[ \bm{M}^{(1)}, \dots, \bm{M}^{(|\mathcal{C}|)} \right]^\top$. 
Next, a low-rank representation of $\bm{M}$ is found using Multiple Discriminant Analysis ~\cite{DudaHartStork01, scikit-learn}.
To perform dimensionality reduction, we compute the solution of the Rayleigh coefficient, which is the ratio of the between class scatter to within class scatter. 
Let $\bm{S}_W$ denote the within-class scatter matrix;
i.e, the prior-weighted sum of within-class covariance matrices. 
Here, $\bm{\mu}^{(c)} \in \mathbb{R}^{|D_o|}$ is the mean of the columns of $\bm{M}^{(c)}$.

The total scatter matrix, $\bm{S}_T$, is the covariance over the dataset, $\bm{M}$, and the between-class scatter matrix is defined as $\bm{S}_B = \bm{S}_T - \bm{S}_W$.
We obtain the eigenvectors, $\bm{w}_i$, of the transformation matrix, $\bm{W}$, which maximize the ratio of between-class scatter to within-class scatter by solving $(\bm{S}_B - \lambda_i\bm{S}_W)\bm{w}_i = 0$. 
Let $\bm{W}_k$ be the reduced-rank representation of $\bm{W}$, with eigenvectors corresponding to the $k = |\mathcal{C}| - 1$ largest eigenvalues. 
The distribution of samples are projected onto the $k$-dimensional subspace spanned by $\bm{W}_k$. For notational simplicity, let $\bm{Z} = \bm{W}_k^\top \bm{M}$ and $\bm{z}_t = \bm{W}_k^\top \bm{y}_t^{D_o}$. 
The posterior density is computed as:
\begin{align}
\log p(&C = c \mid \bm{Y}_{1:t}, \bm{s}_0) \nonumber \\
&=\log p(\bm{Y}_{1:t}, \bm{s}_0 \mid C = c) + \log p(C = c) + \eta \nonumber \\
&= -\frac{1}{2}(\bm{z}_t -  \bm{\mu}_{\bm{Z}^{(c)}})^\top \Sigma_{\bm{Z}}^{-1}(\bm{z}_t - \bm{\mu}_{\bm{Z}^{(c)}}) \nonumber \\
& \hspace{4mm} + \log p(C = c) + \eta. \label{eq:logpost}
\end{align}
After dropping the quadratic term $\bm{z}_t^\top \Sigma_{\bm{Z}}^{-1} \bm{z}_t$  from \ref{eq:logpost}, which is independent of $c$, we get the resulting form,
\begin{equation}
    \log p(C = c \mid \bm{Y}_{1:t}, \bm{s}_0) = \bm{\beta}_c^\top\bm{z}_t + \beta_{c0}.
\end{equation}
where $\bm{\beta}_c = \Sigma_{\bm{Z}}^{-1}\bm{\mu}_{\bm{Z}^{(c)}}$ and $\beta_{c0} = -\frac{1}{2} \bm{\mu}_{\bm{Z}^{(c)}}^\top \Sigma_{\bm{Z}}^{-1} \bm{\mu}_{\bm{Z}^{(c)}} + \log \pi^{(c)}$.
The posterior can now be computed by applying the softmax function,
\begin{align}
    p(C = c \mid \bm{Y}_{1:t}, \bm{s}_0) = \frac{e^{\bm{\beta}_c^\top\bm{z}_t + \beta_{c0}}}{\sum_{j \in C} e^{\bm{\beta}_j^\top\bm{z}_t + \beta_{j0}}}. \label{eq:filter_weights}
\end{align}
We assume a prior proportional to the number of samples in each training set; i.e, $p(C = c) = \pi^{(c)} = \frac{N^{(c)}}{\sum_{j \in C} N^{(j)}}$. 

\begin{figure}[t]
    \centering
    \includegraphics[width=\linewidth]{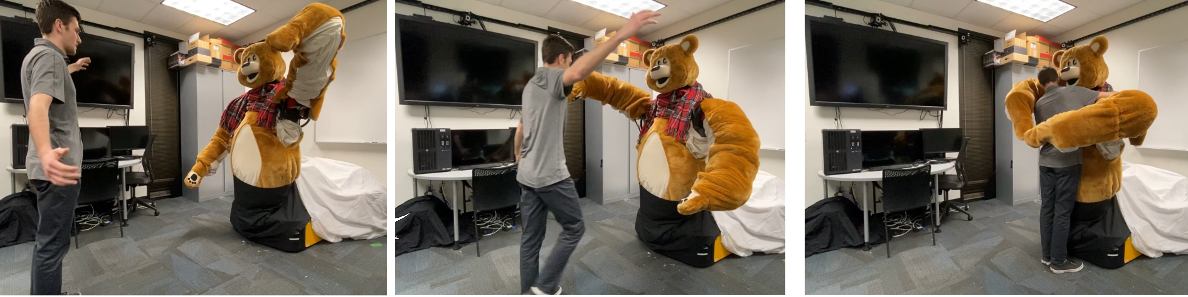}
    \caption{An example of a \textit{left-high} to \textit{right-high} interaction. Left: The participant starts with a left-high interaction. Middle: When switching to the right-high hug, the robot responds accordingly. Right: The participant hugs the robot. }
    \label{fig:exp_sequence}
\end{figure}

\subsection{Interaction Transition}
\label{sec:transition}

When calculating Eq.~\ref{eq:prob_classes} we must be careful to weight the magnitude of the ensemble update with the class probability.
The intuitive reason is that the standard ensemble member update of Eq.~\ref{eq:measurement_update} assumes that the observation was generated from a distribution approximated by that ensemble.
However, if there is only a small probability that the observation was generated from a particular sub-ensemble and we apply a full-magnitude update then we potentially skew the ensemble members with an out-of-distribution measurement.
Thus, unlike Eq.~\ref{eq:measurement_update}, we now weight the gain coefficient with Eq.~\ref{eq:filter_weights}:
\begin{align}
    \boldsymbol{x}^j_{t|t} =& \boldsymbol{x}^j_{t|t-1} +\nonumber\\ &p(C = c \mid \bm{Y}_{1:t}, \bm{s}_0) \boldsymbol{K} (\boldsymbol{\tilde{y}}_{t} - h(\boldsymbol{x}^j_{t|t-1}))
\end{align}
for all $\boldsymbol{x}^j \in \boldsymbol{X}^c$ and all $c \in \mathcal{C}$.

\begin{figure}[t]
    \centering
    \includegraphics[width=\linewidth]{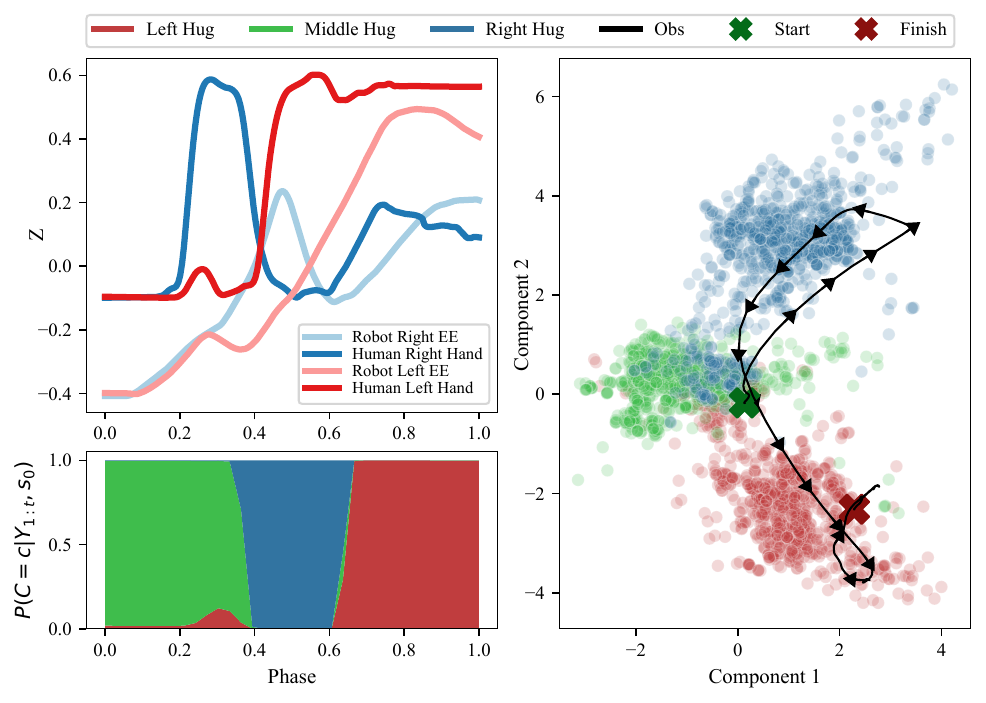}
    \caption{Top left: The observed $z$ positions of the participant's hands and the robot's end effectors during a right-high to left-high interaction. Bottom left: The corresponding interaction class weights for each hug type, with vertical area equal to the class probability. Right: The trajectory of the observed DoF (black arrows) projected to the reduced-rank LDA space, overlaid on the distributions (circles) for each hug type.}
    \label{fig:filter_weights}
\end{figure}

\section{Experimental Design}
\label{sec:exp_setup}
In order to empirically evaluate our algorithm, we conduct an IRB-approved participant study consisting of a hugging scenario between a robot and a human.

\subsection{Training Data Collection}
Motion capture data is collected at $120$ Hz from participants wearing a hat and a wrist band on each hand, totaling three observed modalities. 
$15$ demonstrations of each interaction type (\textit{left-high}, \textit{middle}, and \textit{right-high}) are collected from $15$ different participants, totaling $225$ demonstration hugs per interaction class and $675$ in total. 
The term $\textit{left-high}$ is used to indicate an interaction where the left hand of the robot and the human are raised such that the human's left hand approaches the robot over its right shoulder; the same symmetry holds for the $\textit{right-high}$ interaction, and $\textit{middle}$ is used to denote a hug where the robot hugs with arms at the same height (see Fig. \ref{fig:exp_sequence}). 
We applied a response elicitation technique~\cite{campbell2019learning}, where the robot actuates according to an open loop control policy and the human responds accordingly.
Outlier training demonstrations, which have any DOF value outside of four standard deviations from the distribution of demonstrations at the same point in time, are removed from the dataset. 
After which, an $80-20$ percent train-test split is used to train and validate the methods for mean-squared error (MSE) comparison. 
This results in $439$ training demonstrations and $110$ validation demonstrations.

\subsection{Prediction Methods}
The goal of the study is to compare 1) B-BIP, 2) BIP, 3) Probabilistic Movement Primitives (ProMP)~\cite{ewerton2015learning, maeda2017probabilistic,  paraschos2013probabilistic}, and 4) a LSTM network, all of which are evaluated on $\textit{non-switching}$ interactions (left-high, middle, and right-high hugs) as well as on $\textit{switching}$ interactions (transitions from either left-high to right-high or right-high to left-high hugs). 
The LSTM architecture contains: $28$ hidden units (twice the number of DOFs from the robot), a dropout layer with rate of $0.2$, a batch normalization layer, a fully-connected layer with $28$ units, and a fully-connected layer with $14$ units at the output. 
In the case of ProMP, phase estimation is performed via Dynamic Time Warping (DTW) as described in ~\cite{maeda2017probabilistic}. All methods are trained on the $439$ demonstrations previously described; the alternative methods treat each interaction as one class, while Blending takes into account the interaction class labels.

\subsection{Experimental Hypotheses}
We identified three important factors to be evaluated when comparing the B-BIP to the alternative methods. After every hug, we ask the following questions:
\begin{enumerate}
    \item On a scale of $1$ to $5$, how good was the timing of the robot during the interaction?
    \item On a scale of $1$ to $5$, how well did the robot match your type of hug during the interaction?
    \item On a scale of $1$ to $5$, how responsive was the robot to your motion?
\end{enumerate}

We state our main hypotheses -- which are applied to both non-switching and switching interactions -- as follows:
\begin{itemize}
    \item $\bm{H_1}$: Proposed method better matches the hug type than baseline methods.
    \item $\bm{H_2}$: Proposed method has better timing than baseline methods.
    \item $\bm{H_3}$: Proposed method elicits more responsive behavior than baseline methods.
\end{itemize}

\begin{figure}[t]
    \centering
    \includegraphics[width=\linewidth]{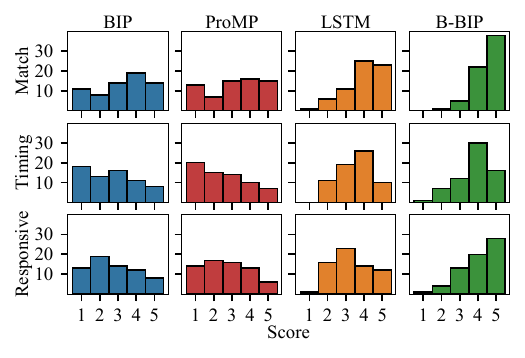}
    \caption{Distribution of scores for the three questions- which are used for hypothesis tests- after switching interactions.}
    \label{fig:histogram}
\end{figure}

\subsection{Participant Study}
The participant study was conducted on $22$ new participants whose data was not used to train the methods, including $4$ female and $18$ male participants between the ages of $18$ and $53$.
A preliminary power analysis for the Wilcoxon test (which we use to conduct our hypotheses, see Section \ref{sec:results}) indicates that a sample size of $n = 20$ is sufficient to achieve a power level of $0.9$, hence the number of participants.
Every participant performed $24$ hugs such that each method was deployed on the three non-switching interactions (totaling $12$) and on switching interactions for the remaining $12$. 
For every hug, the participant is assigned an interaction and the robot is assigned a method, both of which are randomly drawn without replacement.
In the switching interactions, participants performed left-high to right-high hugs, right-high to left-high hugs, and a randomly chosen option from the former two interactions. 
This setup ensures that both switching and non-switching interactions have the same number of samples for every hypothesis test.

\subsection{Quantitative Experimental Design}
\label{sec:exp_quant_design}

In addition to the participant study, offline experiments are conducted to quantitatively evaluate the performance of our proposed algorithm against the baseline methods in terms of MSE. 
All methods predicted a response for the demonstrations in the validation set as well as for demonstrations from a set of unseen switching interactions.
In order to get the ground truth dataset for the switching interactions, we manually design hug trajectories where the robot transitions from left-high to right-high and from right-high to left-high. 
We collect $25$ demonstrations of the left-to-right interaction and $25$ demonstrations of the right-to-left interaction. 
Here, each of the $5$ participants partake in $5$ demonstrations per switching interaction, where the robot executes the designed trajectories.
After which, outliers are removed in the same manner as the validation set, resulting in $39$ demonstrations in total.

\begin{table}[t]
    \begin{minipage}{0.95\linewidth}
        \centering
        \setlength\tabcolsep{4pt}
        \begin{tabular}{cccc}
        Switching & $\bm{H_1}$: $p$-value & $\bm{H_2}$: $p$-value & $\bm{H_3}$: $p$-value \\
        \cmidrule(lr){1-4}\morecmidrules\cmidrule(lr){1-4}
        \multirow{3}{*}{
          \begin{tabular}{r @{ vs. } l}
            B-BIP & BIP \\
            B-BIP & ProMP \\
            B-BIP & LSTM 
          \end{tabular}
          }
         & \sn{3.624324e-07} & \sn{1.942489e-06} & \sn{6.564480e-08} \\
         & \sn{6.159973e-07} & \sn{1.251927e-07} & \sn{2.267898e-07} \\
         & \sn{6.642390e-04} & \cellcolor{gray!20} \sn{9.852936e-02} & \sn{1.824818e-04} \\
         \cmidrule(lr){1-4}
        Supported & \cellcolor{green!15} Yes  & Partially  & \cellcolor{green!15} Yes \\ \\
        Non-Switching & $\bm{H_1}$: $p$-value & $\bm{H_2}$: $p$-value & $\bm{H_3}$: $p$-value \\
        \cmidrule(lr){1-4}\morecmidrules\cmidrule(lr){1-4}
        \multirow{3}{*}{
          \begin{tabular}{r @{ vs. } l}
            B-BIP & BIP \\
            B-BIP & ProMP \\
            B-BIP & LSTM \\
          \end{tabular}}
         & \cellcolor{gray!20} \sn{1.0}  & \cellcolor{gray!20} \sn{0.120512} & \cellcolor{gray!20} \sn{1.0} \\
         & \cellcolor{gray!20} \sn{0.228697} & \cellcolor{white} \sn{0.003080} & \cellcolor{white} \sn{0.015921} \\
         & \cellcolor{white} \sn{0.000480}  & \cellcolor{white} \sn{0.000140} & \cellcolor{white} \sn{0.000057} \\
         \cmidrule(lr){1-4}
         
        Supported & Partially  & Partially  & Partially \\
        \end{tabular}
    \end{minipage}%
    \caption{Top: $p$-values for Switching Interactions. Bottom: $p$-values for Non-Switching Interactions. Grey cells indicate a $p$-value greater than $\alpha=0.05$.\label{tab:survey_significance}}
    
\end{table}

\section{Results and Discussion}
\label{sec:results}
In the following section, we discuss the performance of our proposed method by analyzing participant responses and MSE prediction values with regard to the given hypotheses.
Figure~\ref{fig:exp_sequence} shows an example an interaction where the human switches from a left-high to a right-high hug and the robot reciprocates appropriately.
Additionally, Figure ~\ref{fig:filter_weights} displays predictions from Blending BIP during a right-high to left-high interaction with a test participant.
The top left plot shows the trajectories from the human and robot DoFs during the interaction.
The bottom left plot shows the inferred weights for each interaction, indicating which interaction Blending BIP model estimates is active at every time step.

\subsection{Survey Responses}
\label{sec:results_survey}

Participant survey responses are visualized in Fig.~\ref{fig:histogram} as a histogram, where we can qualitatively observe a difference in distributions. Notably, B-BIP yields the largest number of survey responses with a maximum score of 5 for each question. When performing hypothesis tests, we cannot make the assumption that responses across different treatments are independent; namely, every participant partakes in the same set of methods. 
Additionally, given that the variable of interest, \emph{score}, takes on an integer value from $1$ through $5$ (as in the Likert scale) and that the scores do not appear to be normally distributed, we opt to use a paired non-parametric hypothesis test.
A one-way repeated measures analysis of variance is conducted to test for differences in methods across participant responses by using the Friedman test.
After obtaining a $p$-value of $p < 10^{-5}$, we perform post-hoc analysis by applying the two-sided Wilcoxon signed-rank test to every baseline method paired with our proposed method. To account for multiple comparisons, a Bonferroni correction is applied. This procedure is performed separately for non-switching and switching interaction scores.
For a given hypothesis test, there are $66$ pairs of responses (i.e, $22$ participants who perform $3$ switching/non-switching interactions), and each pair contains the participant's scores to the question of interest for the methods being compared.

The resulting significance results are shown in Table~\ref{tab:survey_significance}.
For switching hugs, we find that participants strongly preferred B-BIP over all baseline methods with respect to hugging type, timing, and responsiveness.
Hypotheses $\bm{H_1}$ and $\bm{H_3}$ were fully supported with participants reporting B-BIP to offer improved timing and responsiveness over BIP, ProMP, and LSTM.
Hypothesis $\bm{H_2}$ was partially supported, with participants finding B-BIP to yield better timing than BIP and ProMP with statistical significance, but not to LSTM.
Responses are more mixed for non-switching hugs, as participants did not prefer B-BIP over BIP in any category, while preferring B-BIP over ProMP only in terms of timing ($\bm{H_2}$) and responsiveness ($\bm{H_3}$).
These results are not unexpected, as any given single interaction falls within the prior distribution modeled by BIP, which means it is not unreasonable for B-BIP to perform similarly.
B-BIP was, however, preferred over the LSTM for all categories, indicating that LSTM struggled to generalize with the small number of training demonstrations.

\begin{figure}[t]
    \centering
    \includegraphics[width=0.8\linewidth]{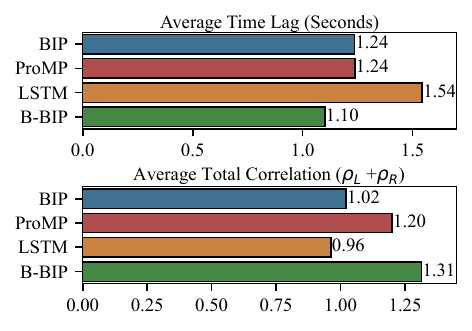}
    \caption{Top: The average of the time lags (in seconds) which maximize the correlation coefficients (below) between human and robot trajectories over all switching interactions from the participant study (see Section~ \ref{sec:responsiveness}). }
    \label{fig:lag_correlation}
\end{figure}

\subsection{Interaction Responsiveness}
\label{sec:responsiveness}
In order to further assess the responsiveness, we evaluate how well the robot matched switching interactions from the group of test participants. 
We calculate the translation (backwards in time, in seconds) which, when applied to the robot end effector trajectories, maximizes the sum of the Pearson correlation coefficients from: 1) the $z$ position of the human's right hand with the $z$ position of the robot's eight end effector, and 2) the $z$ position of the human's left hand with the $z$ position of the robot's left end effector. 
Namely, samples from these matching modalities (from the recorded test interactions) are paired to compute the correlation coefficients, and the average is taken over all switching interactions for every method. 
These calculations take advantage of the symmetrical nature of the interaction, and the intuition here is that preferable methods should exhibit strong correlations with the participant within a small time lag, especially when the switching occurs- with a perfect response having zero time delay. 
Figure~\ref{fig:lag_correlation}, shows that B-BIP has a correlation which is maximized at the smallest time delay of all methods, and B-BIP produces the highest total correlation (i.e, is the sum of the correlations coefficients from the left and right matching modalities), with a value of $1.31$ out of a a maximum value of $2.0$. 
These findings provide further support toward the hypothesis that B-BIP exhibits better responsiveness during switching interactions.

\begin{table}
    \begin{minipage}{0.95\linewidth}
        \centering
        \begin{tabular}{ccc}
             & \thead{Switching} & \thead{Non-Switching} \\
             \cmidrule(lr){1-3}\morecmidrules\cmidrule(lr){1-3}
             BIP          & \cellcolor{white} 0.127 $\pm$ 0.008     & \cellcolor{white!15} 0.040 $\pm$ 0.002 \\
             ProMP        & \cellcolor{white!15} 0.128 $\pm$ 0.005  & \cellcolor{white!20} 0.036 $\pm$ 0.001 \\
             LSTM         & \cellcolor{white} 0.107 $\pm$ 0.005     & \cellcolor{white} 0.102 $\pm$ 0.004\\
             B-BIP & \cellcolor{green!15} 0.062 $\pm$ 0.005   & \cellcolor{green!15} 0.018 $\pm$ 0.000 \\
             \cmidrule(lr){1-3}\morecmidrules\cmidrule(lr){1-3}
        \end{tabular}
    \end{minipage}%
    \caption{MSE values of the controlled DoFs (predicted at each time step) compared to the ground-truth response, using all validation demonstrations. Green cells indicate the method with the smallest mean values. Tukey's Range Test indicates statist\label{tab:mae_results}ical significance for B-BIP in all cases, having $p < 10^{-5}$ in all cases.} 
\end{table}

\subsection{Quantitative Analysis}

The MSE values for all controlled DoFs in the quantitative offline experiments are shown in Table~\ref{tab:mae_results}.
We find that B-BIP yields significantly lower prediction errors than all baseline methods for both switching and non-switching hug types.
From the significance in switching hug types, we can conclude that the interaction detection mechanism works as intended and yields more accurate inference than BIP.
The significance for non-switching hugs is somewhat surprising given the lack of statistical significance in the survey responses for the non-switching hugs in Sec.~\ref{sec:results_survey}, and we conjecture that error values of this magnitude do not always consistently result in noticeable behavioral differences.
It is clear, however, that despite non-switching hugs falling within BIP's prior distribution, the wider distribution results in significantly larger spatio-temporal errors when compared to the per-interaction priors of B-BIP.

\section{Conclusion}

In this paper we present a method for learning and blending human-robot interactions from demonstration. 
A carefully-designed user study is conducted to validate whether our method produces a more responsive, timely, and suitably-matching behavior, all of which are supported through qualitative and quantitative analysis. 
Notably, Blending BIP achieves 1) the highest correlation and lowest response lag with test participants, 2) a nearly twofold reduction in mean-squared prediction error compared to the second-best methods, and 3) statistically significant participant responses on switching interactions for most hypotheses.

\end{document}